\documentclass[10pt,twocolumn,letterpaper]{article}

\usepackage{iccv}              %

\definecolor{iccvblue}{rgb}{0.21,0.49,0.74}
\usepackage[pagebackref,breaklinks,colorlinks,allcolors=iccvblue]{hyperref}

\def\paperID{5177} %
\def\confName{ICCV RIWM}
\def\confYear{2025}

\usepackage{amsfonts}       %
\usepackage{nicefrac}       %
\usepackage{microtype}      %
\usepackage{mathtools}
\usepackage[numbers]{natbib}
\usepackage{lipsum}
\usepackage{array}
\usepackage{epsfig}
\usepackage{graphicx}
\usepackage{float}
\usepackage{wrapfig}
\usepackage{amsmath,amssymb,amsthm}
\usepackage{algorithm,algorithmicx,algpseudocode}
\usepackage{listings}

\usepackage{etoolbox}
\makeatletter
\AfterEndEnvironment{algorithm}{\let\@algcomment\relax}
\AtEndEnvironment{algorithm}{\kern2pt\hrule\relax\vskip3pt\@algcomment}
\let\@algcomment\relax
\newcommand\algcomment[1]{\def\@algcomment{\footnotesize#1}}
\renewcommand\fs@ruled{\def\@fs@cfont{\bfseries}\let\@fs@capt\floatc@ruled
  \def\@fs@pre{\hrule height.8pt depth0pt \kern2pt}%
  \def\@fs@post{}%
  \def\@fs@mid{\kern2pt\hrule\kern2pt}%
  \let\@fs@iftopcapt\iftrue}
\makeatother

\usepackage{bm,xspace}
\usepackage{comment}
\usepackage{multirow}
\usepackage{balance}
\usepackage{url}
\usepackage{booktabs}
\usepackage{etoolbox,siunitx}
\usepackage{calc}
\usepackage{pifont,hologo}
\usepackage{color}
\usepackage{adjustbox}
\usepackage{enumitem}
\usepackage{subcaption}
\usepackage{bbding}  %
\usepackage[normalem]{ulem}  %
\PassOptionsToPackage{table}{xcolor}
\usepackage{colortbl}
\usepackage{xcolor}         %
\usepackage[table]{xcolor}
\usepackage{graphicx}

\newcommand{\sexyname}{\textsc{Aether}}

\definecolor{myblue}{HTML}{118ab2}
\definecolor{red}{HTML}{ef476f}
\definecolor{orange}{HTML}{cc7700}
\definecolor{mygray}{HTML}{efefef}
\definecolor{darkgreen}{HTML}{228B22}
\definecolor{mydarkgray}{HTML}{757575}

\newcommand{\figref}[1]{Fig.~\ref{#1}}
\newcommand{\tabref}[1]{Tab.~\ref{#1}}
\newcommand{\secref}[1]{Sec.~\ref{#1}}
\renewcommand{\eqref}[1]{Eq.~\ref{#1}}

\newcolumntype{x}[1]{>{\centering\arraybackslash}p{#1}}
\newcolumntype{y}[1]{>{\raggedright\arraybackslash}p{#1}}
\newcolumntype{z}[1]{>{\raggedleft\arraybackslash}p{#1}}
\newcommand{\tablestyle}[2]{\setlength{\tabcolsep}{#1}\renewcommand{\arraystretch}{#2}\centering\footnotesize}

\DeclareMathSymbol{@}{\mathord}{letters}{"3B}

\makeatletter
\DeclareRobustCommand\onedot{\futurelet\@let@token\@onedot}
\def\@onedot{\ifx\@let@token.\else.\null\fi\xspace}

\newcommand*{\Rom}[1]{\expandafter\@slowromancap\romannumeral #1@}
\newcommand*{\rom}[1]{\expandafter\romannumeral #1}

\def\1{\bm{1}}

\let\originalleft\left
\let\originalright\right
\renewcommand{\left}{\mathopen{}\mathclose\bgroup\originalleft}
\renewcommand{\right}{\aftergroup\egroup\originalright}

\definecolor{resnet}{HTML}{264653}
\definecolor{r3m}{HTML}{264653}
\definecolor{vit}{HTML}{2A9D8F}
\definecolor{vc1}{HTML}{2A9D8F}
\definecolor{multivit}{HTML}{E9C46A}
\definecolor{multimae}{HTML}{E9C46A}
\definecolor{spunet}{HTML}{F4A261}
\definecolor{ponderv2}{HTML}{F4A261}
\definecolor{pointnet}{HTML}{E76F51}
\definecolor{cyan}{HTML}{264653}

\usepackage{pifont}%

\title{\sexyname{}: Geometric-Aware Unified World Modeling}

\author{
  Haoyi Zhu$^*$\quad Yifan Wang$^*$\quad Jianjun Zhou$^*$\quad Wenzheng Chang$^*$\quad Yang Zhou$^*$\\ Zizun Li$^*$\quad
  Junyi Chen$^*$ \quad
  Chunhua Shen \quad Jiangmiao Pang \quad Tong He$^{\textsuperscript{\dag}}$ \\
  USTC\quad Shanghai AI Lab\quad SII\quad SJTU\quad ZJU\quad FDU\\
  $^{*}$Equal Contribution \quad $^{\textsuperscript{\dag}}$Corresponding Author\\
  \url{https://aether-world.github.io/}
}

\begin{document}

\twocolumn[{
\maketitle
\thispagestyle{empty}
\begin{center}
    \centering
    \includegraphics[width=\textwidth]{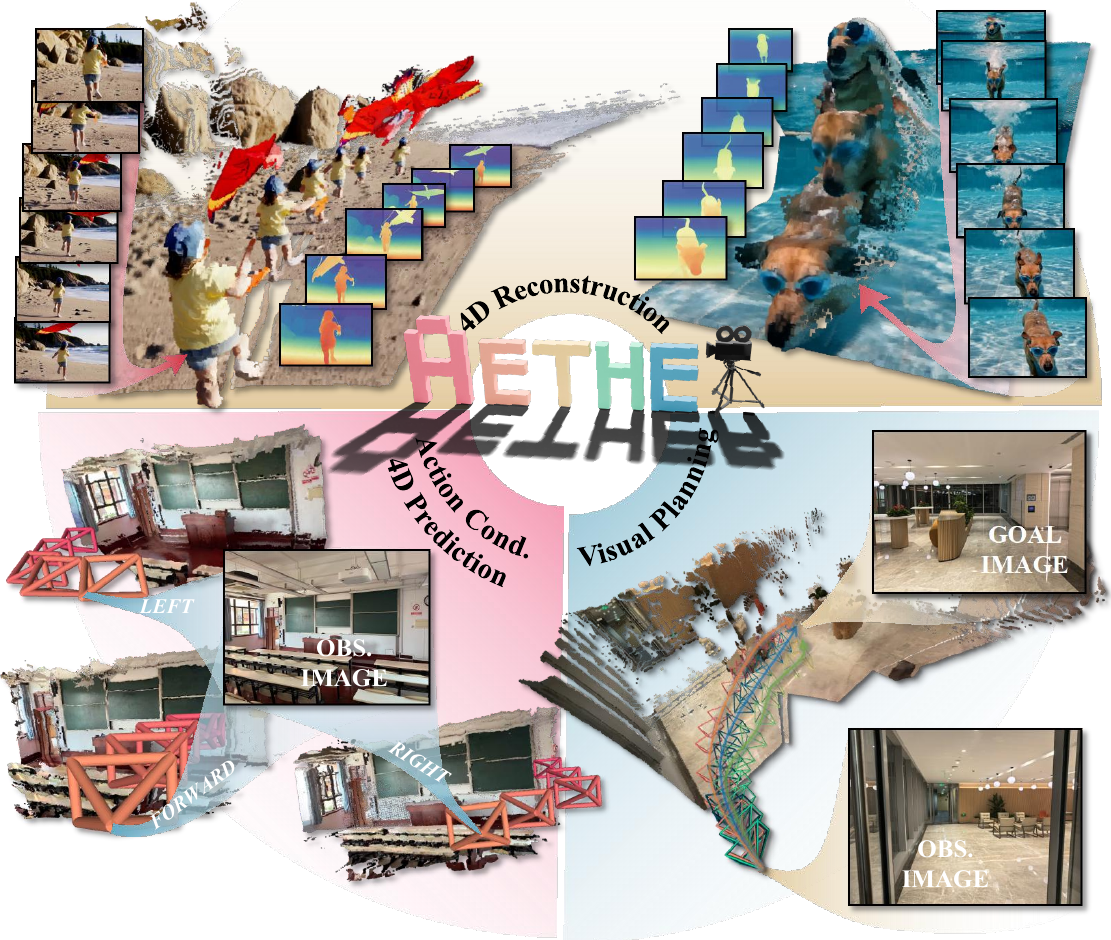}
    \captionof{figure}{\textbf{An overview of \sexyname{}, trained entirely on synthetic data.} The figure highlights its three key capabilities: 4D reconstruction, action-conditioned 4D prediction, and visual planning, all demonstrated on unseen real-world data. The 4D reconstruction examples are derived from MovieGen~\cite{polyak2025moviegencastmedia} and Veo 2~\cite{veo2} generated videos, while the action-conditioned prediction uses an observation image from a university classroom. The visual planning example utilizes observation and goal images from an office building. Better viewed when zoomed in. Additional visualizations can be found on our website.}
    \label{fig:teaser}
\end{center}
\vspace{1.5em}
}
]

\begin{abstract}
The integration of geometric reconstruction and generative modeling remains a critical challenge in developing AI systems capable of human-like spatial reasoning. This paper proposes \sexyname{}, a unified framework that enables geometry-aware reasoning in world models by jointly optimizing three core capabilities: (1) 4D dynamic reconstruction, (2) action-conditioned video prediction, and (3) goal-conditioned visual planning. Through task-interleaved feature learning, \sexyname{} achieves synergistic knowledge sharing across reconstruction, prediction, and planning objectives. Building upon video generation models, our framework demonstrates zero-shot synthetic-to-real generalization despite never observing real-world data during training. Furthermore, our approach achieves zero-shot generalization in both action following and reconstruction tasks, thanks to its intrinsic geometric modeling. Notably, even without real-world data, its reconstruction performance is comparable with or even better than that of domain-specific models. Additionally, \sexyname{} employs camera trajectories as geometry-informed action spaces, enabling effective action-conditioned prediction and visual planning.
We hope our work inspires the community to explore new frontiers in physically-reasonable world modeling and its applications.

\end{abstract}

\section{Introduction}
\label{sec:introduction}

\emph{``Prediction is not just one of the things your brain does. It is the primary function of the neocortex.''}\\
\hspace*{\fill} — Jeff Hawkins, \textit{On Intelligence} (2004)
\vspace{0.5em}

The development of visual intelligence systems capable of comprehending and forecasting the physical world remains a cornerstone of AI research.
World models have emerged as a foundational paradigm for building autonomous systems that not only perceive but also anticipate environmental dynamics to make reasonable actions. At their core, three capabilities stand out: First, perception equips the system with the ability to capture the intricate four-dimensional (4D) changes—integrating spatial and temporal information—that are essential for understanding the physical world~\cite{zhao2022particlesfm,wang20243d,wang2024dust3r,wang2025continuous,leroy2024grounding,zhang2024monst3r}. This continuous sensing of dynamic cues enables a geometric representation of the environment. Second, prediction leverages this perceptual information to forecast how the environment will evolve under specific actions, thereby providing a foresight into future states~\cite{hong2022cogvideo,yang2024cogvideox,blattmann2023stable,kong2024hunyuanvideo,jin2024pyramidal,HaCohen2024LTXVideo,wan2.1}. Finally, planning uses these predictive insights to determine the optimal sequence of actions required to achieve a given goal. Together, these three aspects empower world models to not only represent the current state of the environment but also to anticipate and navigate its future dynamics effectively.

Motivated by these principles, we introduce \sexyname{}, a unified framework that, for the first time, bridges reconstruction, prediction, and planning, as shown in \figref{fig:teaser}. \sexyname{} leverages pre-trained video generation models~\cite{hong2022cogvideo,yang2024cogvideox} and is further refined via post-training with synthetic 4D data. Although multiple action modalities exist, ranging from keyboard inputs~\cite{erez2012infinite,parkerholder2024genie2,alonso2025diffusiondiamond,yu2024gamefactory,oasis2024} to human or robotic motions~\cite{zhang2024mimicmotionhighqualityhumanmotion,zhu2024point,zhu2024spa,fang2023rh20t} and point flows~\cite{wang2024motionctrl,geng2024motionpromptingcontrollingvideo}, we choose camera pose trajectories as our global action representation. This choice is particularly effective for ego-view tasks: in navigation, camera trajectories directly correspond to the navigation paths, while in robotic manipulation, the movement of an in-hand camera captures the 6D motion of the end effector. 
To address the scarcity of 4D data, we utilize RGB-D synthetic video data and propose a robust camera pose annotation pipeline to reconstruct full 4D dynamics.

\begin{figure*}[!t]
    \centering
    \includegraphics[width=0.9\linewidth]{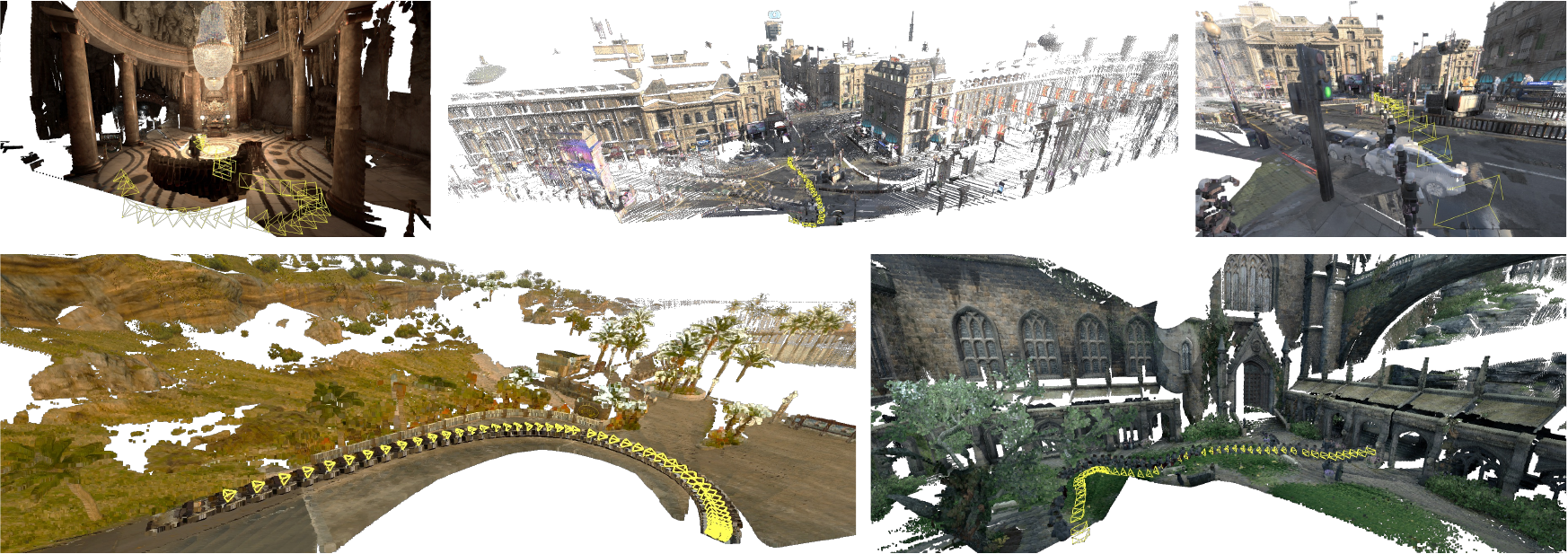}
    \vspace{-1em}
    \caption{Some visualization results of data annotated through our pipeline. Better viewed when zoomed in.}
    \vspace{-1.5em}
    \label{fig:data_example}
\end{figure*}

Through a simple training strategy that randomly combines input and output modalities, our method transforms the base video generation model into a unified, multi-task world model with three key capabilities: (1) Depth and camera pose estimation from full video sequences; (2) Video prediction conditioned on an initial observation—with the option to incorporate a camera trajectory action; and (3) Goal-conditioned visual planning based on observation–goal image pairs.
We transform depth videos into scale-invariant normalized disparity representations to meet the tokenization requirements of video VAEs. Simultaneously, we encode camera trajectories as scale-invariant raymap sequence representations, structured to align with the spatiotemporal framework of diffusion transformers (DiTs). By dynamically integrating cross-task and cross-modal conditioning signals during training, our framework enables synergistic knowledge transfer across heterogeneous inputs, facilitating joint optimization for multi-task generative modeling.

In summary, this work introduces \sexyname{}, a unified world model that integrates reconstruction, prediction, and planning through multi-task learning on synthetic 4D data. We propose a robust automatic data annotation pipeline to obtain accurate 4D geometry knowledge. By combining geometric reasoning with generative priors, our framework achieves robust zero-shot transfer to real-world tasks, demonstrating accuracy comparable to SOTA reconstruction models while enabling actionable planning capabilities. The results underscore the value of synergistic 4D modeling for advancing spatial intelligence in AI systems. We hope that \sexyname{} will serve as an effective starter framework for the community to explore post-training world models with scalable synthetic data.

\section{4D Synthetic Data Annotation Pipeline}
\label{sec:data-pipeline}

\begin{figure}[!t]
    \centering
    \includegraphics[width=0.95\linewidth]{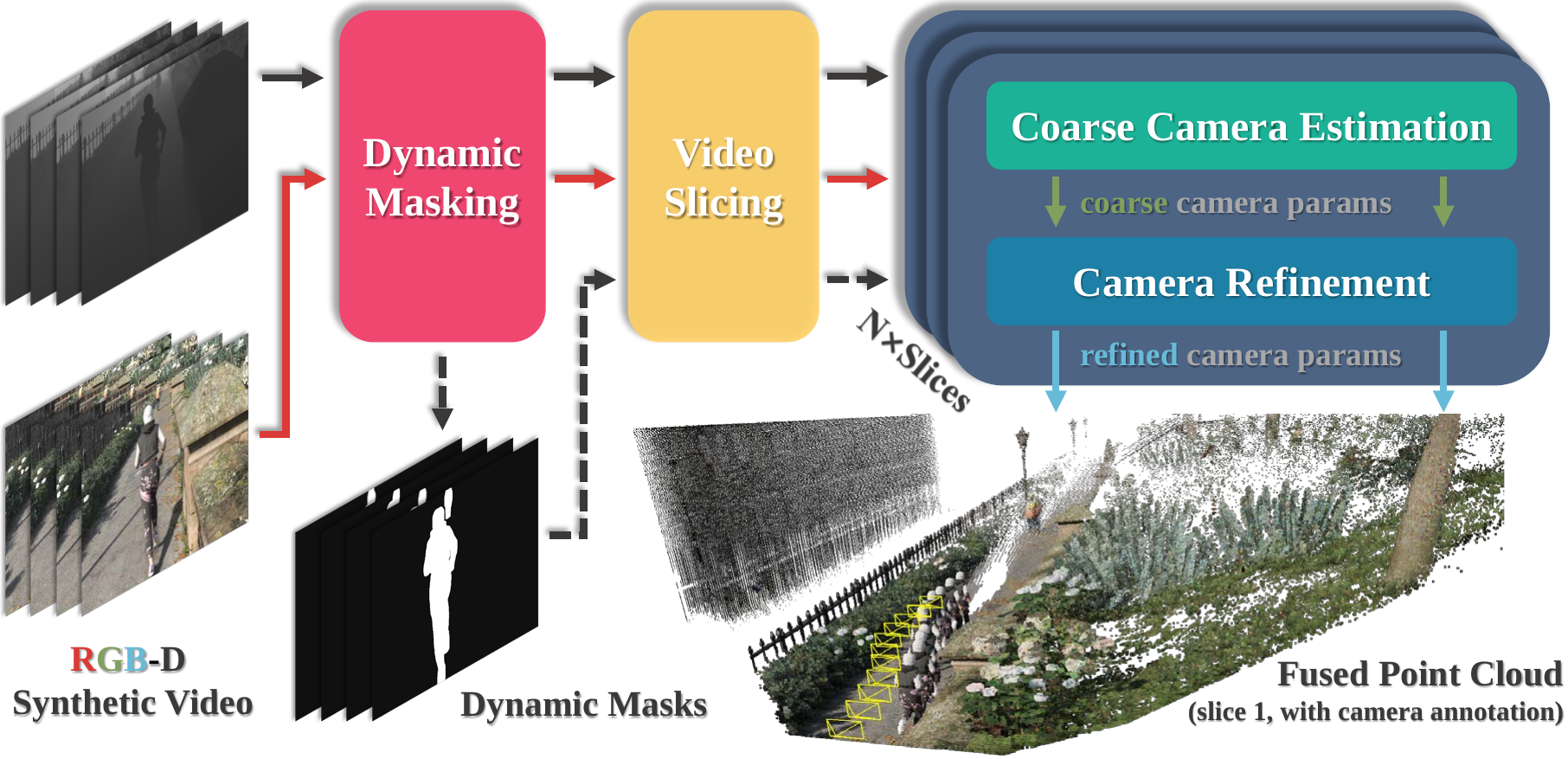}
    \vspace{-1em}
    \caption{Our robust automatic camera annotation pipeline.}
    \vspace{-2.2em}
    \label{fig:data_pipeline}
\end{figure}

For the synthetic data source, we follow DA-V~\cite{yang2024depth} and TheMatrix~\cite{feng2024matrix} to collect large-scale synthetic data with high-quality video depth data.
With high-resolution RGB videos and corresponding per-frame depth maps collected, we built a \textbf{robust} and fully \textbf{automatic} camera annotation pipeline for both camera extrinsics and intrinsics.
As illustrated in \figref{fig:data_pipeline}, the pipeline has four stages: (1) object-level dynamic masking, (2) reconstruction-friendly video slicing, (3) coarse camera localization and calibration, and (4) tracking-based camera refinement with bundle adjustment. We present several visualizations of our annotated data in \figref{fig:data_example}, ranging from indoor to outdoor scenes, and from static to dynamic scenarios, demonstrating the robustness and accuracy of our annotation method.

\noindent\textbf{Dynamic Masking.} 
Distinguishing between dynamic and static regions is crucial for accurate camera parameters estimation. 
Here, we utilize semantic categories that are potentially dynamic (e.g., cars, people) to segment dynamic objects. Although this may occasionally misclassify static objects, such as stationary parked cars, as dynamic, we find it more robust than flow-based segmentation methods. Specifically, we use Grounded SAM 2 ~\cite{ren2024grounded} to ensure the temporal consistency of dynamic masks over long sequences.

\noindent\textbf{Video Slicing.}
Video slicing plays a critical role in 3D reconstruction by serving two key purposes: First, it eliminates unsuitable video segments (such as scene cuts or motion-blurred frames) that could compromise reconstruction quality. Second, it segments long videos into shorter, temporally coherent clips to enhance processing efficiency. The specific criteria for frame removal are as follows: (1) \textit{Insufficient Feature Points}: We employ the SIFT~\cite{lowe2004distinctive} feature descriptor to extract keypoints from each frame. Frames exhibiting insufficient SIFT keypoints are discarded to ensure robust correspondence estimation. Additionally, frames containing regions with insufficient texture due to low illumination are excluded, as such areas typically exhibit poor feature discriminability and pose challenges for reliable matching. (2) \textit{Large Areas of Dynamic Regions}: Frames where dynamic regions (obtained from dynamic annotation) dominate over static regions can introduce ambiguity in camera pose estimation. Such frames are filtered out to ensure robust results. (3) \textit{Large Motion or Inaccurate Correspondence}: Using an off-the-shelf optical flow estimator, RAFT~\cite{teed2020raft}, we estimate the magnitude of motion. If these magnitudes exceed a predefined threshold, we truncate the sequence at the current frame, retaining all preceding frames as a valid segment. Similarly, if the ratio of forward-to-backward optical flow errors surpasses a threshold value, we truncate the current frames to ensure temporal coherence.

\noindent\textbf{Coarse Camera Estimation.}
For each video slice, we first use DroidCalib ~\cite{hagemann2023deep} to perform a coarse estimation of the camera parameters, leveraging the depth information from static regions. However, due to the lower input resolution of the DroidCalib model and the limited accuracy of its correspondence estimation, a refinement process is necessary to obtain precise camera parameters.

\noindent\textbf{Camera Refinement.} We begin camera refinement by employing the state-of-the-art tracker, CoTracker3 ~\cite{karaev2024cotracker3}, to capture accurate long-term correspondences across the entire slice. SIFT ~\cite{lowe2004distinctive} and SuperPoint ~\cite{detone2018superpoint} feature points are extracted from static regions, and then tracked to form correspondences. Subsequently, bundle adjustment is performed on all frames to minimize the accumulated reprojection error of all correspondences. With access to high-quality dense depth, we apply forward-backward reprojection to estimate and minimize errors in 3D space ~\cite{chen2019self}, which improves per-frame camera accuracy while preserving inter-frame geometric consistency. Specifically, we solve the nonlinear optimization problem by Ceres Solver ~\cite{agarwal2012ceres}, and the Cauchy loss function is applied to measure correspondence residuals, which accounts for the problem's sparsity.

\begin{figure*}[!t]
    \centering
    \includegraphics[width=0.9\linewidth]{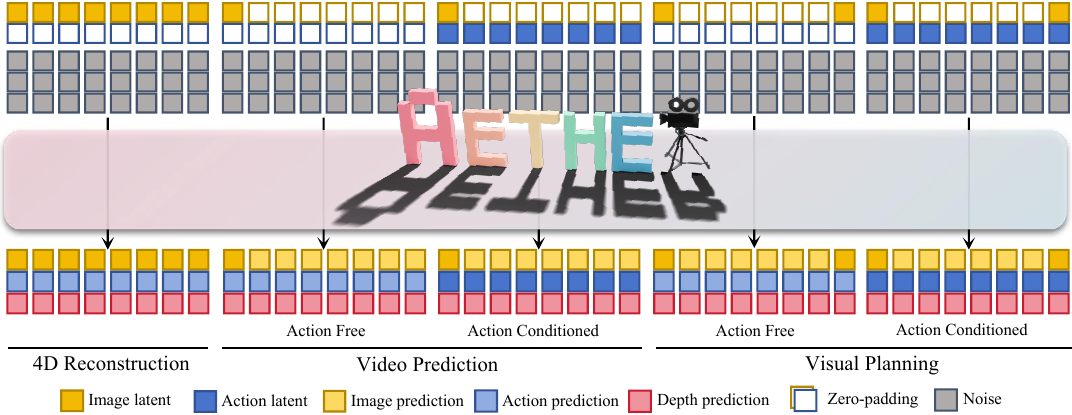}
    \vspace{-0.9em}
    \caption{The overall pipeline of \sexyname{}. With different condition combinations, \sexyname{} can serve different tasks.}
    \vspace{-1.5em}
    \label{fig:model_pipeline}
\end{figure*}

\section{\sexyname{} Multi-Task World Model}
\label{sec:model}

In this section, we introduce how we post-train a base video diffusion model into a unified multi-task world model \sexyname{}. We use \href{https://huggingface.co/THUDM/CogVideoX-5b-I2V}{CogVideoX-5b-I2V}~\cite{yang2024cogvideox} as our base model. We first give an overview of our framework in \secref{sec:model-method-overview}, then we detail on the input process of depth videos and camera pose trajectories in \secref{sec:model-depth-video-process} and \secref{sec:model-camera-pose-process}. Finally, we show how we do model training in \secref{sec:model-training}.

\subsection{Method Overview}
\label{sec:model-method-overview}

Mainstream video diffusion models~\cite{ho2020denoising,lu2022dpm} typically involve two processes: a forward (noising) process and a reverse (denoising) process. The forward process incrementally adds Gaussian noise, denoted as $\epsilon \sim \mathcal{N}(0, \mathbf{I})$, to a clean latent sample $\mathbf{z}_0 \in \mathbb{R}^{k \times c \times h \times w}$, where $k, c, h, w$ represent the dimensions of the video latents. Through this process, the clean $\mathbf{z}_0$ is gradually transformed into a noisy latent $\mathbf{z}_t$. In the reverse process, a learned denoising model $\epsilon_\theta$ progressively removes the noise from $\mathbf{z}_t$ to reconstruct the original latent representation. The denoising model $\epsilon_\theta$ is conditioned on auxiliary inputs $\mathbf{c}$ and the diffusion timestep $t$.

In our method, the target latent $\mathbf{z}_0$ comprises three modalities: color video latents ${\mathbf{z_{\mathrm{c}}}}_0$, depth video latents $\mathbf{z_{\mathrm{d}}}_0$, and action latents $\mathbf{z_{\mathrm{a}}}_0$. The model additionally takes two types of conditions as input: color video conditions $\mathbf{c}_{\mathrm{c}}$ and action conditions $\mathbf{c}_{\mathrm{a}}$. For the action modality, we choose \textit{camera pose trajectory} as a global action, facilitated by our automated camera pose annotation pipeline described earlier. All latents and conditions are channel-wise concatenated. 
The training objective of \sexyname{} can be expressed as:

\begin{equation}
\mathcal{L}_\theta = \mathbb{E}_{\substack{\epsilon \sim \mathcal{N}(0,\mathbf{I}) \\ t \sim \mathcal{U}(1,\mathcal{T}) \\ \mathbf{z}_0 = \mathbf{z_{\mathrm{c}}}_0 \otimes \mathbf{z_{\mathrm{d}}}_0 \otimes \mathbf{z_{\mathrm{a}}}_0 \\ \mathbf{c} = \mathbf{c}_{\mathrm{c}} \otimes \mathbf{c}_{\mathrm{a}}}} \| \epsilon - \epsilon_\theta(\mathbf{z}_t, t, \mathbf{c}) \|^2,
\end{equation}
where $\otimes$ denotes the channel-wise concatenation operation, $\mathcal{U}(\cdot)$ represents a uniform distribution, and $\mathcal{T}$ denotes the denoising steps.

The multi-task objective of \sexyname{} is determined by the specific conditions $\mathbf{c}$ for different tasks. 
(1) \textit{Reconstruction}: $\mathbf{c}_{\mathrm{c}}$ represents the input video latents.
(2) \textit{Video prediction}: $\mathbf{c}_{\mathrm{c}}$ takes the latent of observation image as the first frame, while other latents are zero-masked.
(3) \textit{Goal-conditioned visual planning}: The first and last latents of $\mathbf{c}_{\mathrm{c}}$ correspond to the observation and goal images, respectively, with all intermediate latents zero-padded.
For the action condition $\mathbf{c}_{\mathrm{a}}$, it is either entirely zero-masked or contains the full target camera pose trajectory in action-free or action-conditioned control cases. Illustrations are show in \figref{fig:model_pipeline}.

\subsection{Depth Videos Process}
\label{sec:model-depth-video-process}

Given a depth video $\mathbf{x}_d$, we first clip the depth values to a predefined range $[d_\mathrm{min}, d_\mathrm{max}]$. Next, we apply a square root transformation and subsequently compute the reciprocal to convert the depth values into disparity, as described in~\cite{song2025depthmaster}. Each disparity video clip is then normalized in a scale-invariant manner. Subsequently, the normalized disparity values are linearly mapped from $[0, 1]$ to $[-1, 1]$. To meet the input requirements of the VAE, the single-channel disparity map is replicated across three channels, as done in prior works~\cite{ke2024repurposing,yang2024depth}. The final depth latent is computed as:
\begin{align}
    \mathbf{x}_\mathrm{disp} &= \frac{1}{\sqrt{\operatorname{clip}(\mathbf{x}_d, d_\mathrm{min}, d_\mathrm{max})}}, \\
    \hat{\mathbf{x}}_\mathrm{disp} &= \frac{\mathbf{x}_\mathrm{disp}}{\max{(\mathbf{x}_\mathrm{disp})}}\times 2 - 1, \\
    \mathbf{z}_d &= \mathcal{E}\left(\hat{\mathbf{x}}_\mathrm{disp} \otimes \mathbf{1}_3\right),
\end{align}
where $\mathcal{E}$ denotes the 3D VAE, and $\otimes \mathbf{1}_3$ represents the channel-wise replication of 3 times. The above operations are designed to be compatible with the pretrained 3D VAE model, ensuring minimal reconstruction error.

\subsection{Camera Trajectories Process} 
\label{sec:model-camera-pose-process}

We transform camera parameters into raymap videos \cite{chen2024iiseeautoregressive} so that video diffusion can process them compatibly. Specifically, given the intrinsic matrix $\mathbf{K} \in \mathbb{R}^{T \times 3 \times 3}$ and the extrinsic matrix $\mathbf{E} \in \mathbb{R}^{T \times 4 \times 4}$, the transformation process can be described as follows.

\noindent \textbf{Translation Scaling and Normalization.}
The translation component of the camera pose (inverse of extrinsic matrix), $\mathbf{t} \in \mathbb{R}^3$, is first scaled by a constant factor $s_\mathrm{ray}$ and normalized using the maximum disparity value $d_\mathrm{max}$. To suppress large values, we then pass it through a signed $\log(1 + \cdot)$ transformation:
\begin{align}
    \mathbf{t}' &= \frac{\mathbf{t}}{\max{(x_{\mathrm{disp}})}} \cdot s_\mathrm{ray},\\
    \mathbf{t}_\mathrm{log} &= \operatorname{sign}(\mathbf{t}') \cdot \log(1 + |\mathbf{t}'|),
\end{align}
where $s_\mathrm{ray}$ is a predefined scaling factor.

\noindent \textbf{Raymap Construction.}
Using the intrinsic matrix $\mathbf{K}$, we compute the camera ray directions $\mathbf{r}_\mathrm{d}$ in homogeneous coordinates for each pixel. Note that we do \textit{not} unit normalize it but let it have a unit value along z axis. The ray origins $\mathbf{r}_\mathrm{o}$ are set to the translation $\mathbf{t}_\mathrm{log}$, replicated across the spatial dimensions. The raymap in the world coordinate system is obtained by transforming the ray directions $\mathbf{r}_\mathrm{d}$ using the extrinsic matrix $\mathbf{E}$. The final raymap $\mathbf{r}$ consists of 6 channels: 3 for the ray directions $\mathbf{r}_\mathrm{d}$ and 3 for the ray origins $\mathbf{r}_\mathrm{o}$.

\noindent \textbf{Resolution Downsampling.}
To align the raymap with the latent feature dimensions from the VAE, we perform adjustments both spatially and temporally. Spatially, the raymap is downsampled by a factor of 8 using bilinear interpolation. Temporally, every consecutive group of 4 frames is concatenated along the channel dimension. The resulting rearranged tensor is denoted as $\mathbf{z}_a$

\noindent \textbf{Converting raymap back to camera matrix.}
Given generated raymap sequences rearranged by the time axis $\hat{\mathbf{r}} \in \mathbb{R}^{T \times 6 \times h \times w} = [\hat{\mathbf{r}_\mathrm{d}},\hat{\mathbf{r}_\mathrm{o}}]$, we first recover the ray origins by:
\begin{equation}
    \hat{\mathbf{r}_\mathrm{o}}' = \frac{1}{s_\mathrm{ray}}\cdot \operatorname{sign}(\hat{\mathbf{r}_\mathrm{o}}) \cdot \left(\exp(|\hat{\mathbf{r}_\mathrm{o}}|) - 1\right),
\end{equation}
Then, we can recover both the intrinsics and extrinsics through Alg.~\ref{alg:raymap-to-cam-pose} in the supplementary material.

\subsection{Model Training}
\label{sec:model-training}
We initialize \sexyname{} with pre-trained CogVideo-5b-I2V~\cite{yang2024cogvideox} weights, excluding the additional input and output projection layer channels for depth and raymap action trajectories, which are initialized to zero. Since text prompt conditions are not used, an empty text embedding is provided during both training and inference.

As the dataset we use contains video clips with variable lengths and frames per second (FPS), we randomly select $T \in \{17, 25, 33, 41\}$ frames, and the FPS is randomly sampled from $\{8, 10, 12, 15, 24\}$. The RoPE~\cite{su2024roformer} coefficients are linearly interpolated to align with them.

During training, conditional inputs are randomly masked to generalize across tasks. For $\mathbf{c}_c$, masking probabilities are: 30\% for both observation and goal images (visual planning tasks), 40\% for observation images only (video prediction), 28\% for full-color video latents (4D reconstruction), and 2\% for masking all of $\mathbf{c}_c$. For $\mathbf{c}_a$, trajectory latents are either kept or fully masked with equal probability (supporting action-free or action-conditioned tasks with raymap conditions). This strategy enables the model to adapt to diverse tasks and input condition settings.

Our training process consists of two stages. In the first stage, we adopt the loss function of a standard latent diffusion model, which minimizes the mean squared error (MSE) in the latent space. In the second stage, we refine the generated outputs by decoding them into the image space. Specifically, we introduce three additional loss terms: a Multi-Scale Structure Similarity (MS-SSIM) loss~\cite{wang2003multiscale} for color video, a scale- and shift-invariant loss~\cite{ranftl2020towards} for depth videos, and a scale- and shift-invariant pointmap loss~\cite{wang2024dust3r} for pointmaps projected from the generated depths and raymaps. 
Further details on the stage 2 loss functions are provided in the supplementary material. 
Notably, the second stage takes about $\frac{1}{4}$ of the training steps used in the first stage.

We employ a hybrid training strategy combining Fully Sharded Data Parallel (FSDP)~\cite{zhao2023pytorch} with Zero-2 optimization within compute nodes and Distributed Data Parallel (DDP) across nodes. Since depth videos require online normalization, the VAE encoder is also run online during training and operates under DDP. Our implementation processes a local batch size of 4 per GPU, resulting in an effective batch size of 320 samples across 80 A100-80GB GPUs. Training is conducted over two weeks using the AdamW~\cite{loshchilov2017decoupled} optimizer with a OneCycle~\cite{smith2019super} learning rate scheduler.

\section{Reconstruction Experiments}
\label{sec:recon_exp}

\begin{table*}[!tb]\centering
\caption{\textbf{Video depth Evaluation.} Methods requiring global alignment are marked ``GA".}\label{tab:video-depth}
\vspace{-1.0em}
\tablestyle{15.5pt}{0.95}
\begin{tabular}{l|cc|cc|cc}
\toprule
\multirow{2}{*}{Method} &\multicolumn{2}{c|}{Sintel~\cite{butler2012naturalistic}} &\multicolumn{2}{c|}{BONN~\cite{palazzolo2019refusion}} &\multicolumn{2}{c}{KITTI~\cite{geiger2013vision}} \\\cmidrule(lr){2-3}\cmidrule(lr){4-5}\cmidrule(lr){6-7}
&Abs Rel~$\downarrow$ &$\delta<1.25~\uparrow$ &Abs Rel~$\downarrow$ &$\delta<1.25~\uparrow$ &Abs Rel~$\downarrow$ &$\delta<1.25~\uparrow$ \\\midrule
\multicolumn{7}{l}{\textit{Reconstruction Methods. Alignment: per-sequence scale}} \vspace{0.25em}\\
DUSt3R-GA~\cite{wang2024dust3r} &0.656 &45.2 &0.155 &83.3 &0.144 &81.3 \\
MASt3R-GA~\cite{leroy2024grounding} &0.641 &43.9 &0.252 &70.1 &0.183 &74.5 \\
MonST3R-GA~\cite{zhang2024monst3r} &\underline{0.378} &\textbf{55.8} &\textbf{0.067} &\textbf{96.3} &0.168 &74.4 \\
Spann3R~\cite{wang20243d} &0.622 &42.6 &0.144 &81.3 &0.198 &73.7 \\
CUT3R~\cite{wang2025continuous} &0.421 &47.9 &\underline{0.078} &\underline{93.7} &\underline{0.118} &\underline{88.1} \\
\textbf{\sexyname{} (Ours)} &\textbf{0.324} &\underline{50.2} &0.273 &59.4 &\textbf{0.056} &\textbf{97.8} \\\midrule
\multicolumn{7}{l}{\textit{Diffusion-Based Methods. Alignment: per-sequence scale\&shift}} \vspace{0.25em}\\
ChronoDepth~\cite{shao2024learning} &\underline{0.429} &38.3 &0.318 &51.8 &0.252 &54.3 \\
DepthCrafter~\cite{hu2024depthcrafter} &0.590 &\underline{55.5} &\textbf{0.253} &\underline{56.3} &0.124 &86.5 \\
DA-V~\cite{yang2024depth} &1.252 &43.7 &0.457 &31.1 &\underline{0.094} &\underline{93.0} \\
\textbf{\sexyname{} (Ours)} &\textbf{0.314} &\textbf{60.4} &\underline{0.308} &\textbf{60.2} &\textbf{0.054} &\textbf{97.7}\\
\bottomrule
\end{tabular}
\vspace{-1em}
\end{table*}

\begin{table*}[!tp]\centering
\caption{\textbf{Evaluation on Camera Pose Estimation.}}\label{tab:camera-pose}
\vspace{-1.0em}
\tablestyle{8.1pt}{0.95}
\begin{tabular}{l|ccc|ccc|ccc}\toprule
\multirow{2}{*}{Method} &\multicolumn{3}{c|}{Sintel~\cite{butler2012naturalistic}} &\multicolumn{3}{c|}{TUM-dynamics~\cite{sturm2012benchmark}} &\multicolumn{3}{c}{ScanNet~\cite{dai2017scannet}} \\\cmidrule{2-10}
&ATE~$\downarrow$ &RPE trans~$\downarrow$ &RPE rot~$\downarrow$ &ATE~$\downarrow$ &RPE trans~$\downarrow$ &RPE rot~$\downarrow$ &ATE~$\downarrow$ &RPE trans~$\downarrow$ &RPE rot~$\downarrow$ \\\midrule
\multicolumn{10}{l}{\textit{Optimization-based Methods}} \vspace{0.25em}\\
Particle-SfM~\cite{zhao2022particlesfm} &\underline{0.129} &\textbf{0.031} &\textbf{0.535} &- &- &- &0.136 &0.023 &0.836 \\
Robust-CVD~\cite{kopf2021robust} &0.360 &0.154 &3.443 &0.153 &0.026 &3.528 &0.227 &0.064 &7.374 \\
CasualSAM~\cite{zhang2022structure} &0.141 &\underline{0.035} &\underline{0.615} &\underline{0.071} &\textbf{0.010}&1.712 &0.158 &0.034 &1.618 \\
DUSt3R-GA~\cite{wang2024dust3r} &0.417 &0.250 &5.796 &0.083 &0.017 &3.567 &0.081 &0.028 &0.784 \\
MASt3R-GA~\cite{leroy2024grounding} &0.185 &0.060 &1.496 &\textbf{0.038} &\underline{0.012} &\textbf{0.448} &\underline{0.078} &\underline{0.020} &\textbf{0.475} \\
MonST3R-GA~\cite{zhang2024monst3r} &\textbf{0.111} &0.044 &0.896 &0.098 &0.019 &\underline{0.935} &\textbf{0.077} &\textbf{0.018} &\underline{0.529} \\\midrule
\multicolumn{10}{l}{\textit{Feed-forward Methods}} \vspace{0.25em}\\
DUSt3R~\cite{wang2024dust3r} &0.290 &0.132 &7.869 &0.140 &0.106 &3.286 &0.246 &0.108 &8.210 \\
Spann3R~\cite{wang20243d} &0.329 &0.110 &4.471 &0.056 &0.021 &\underline{0.591} &\textbf{0.096} &\underline{0.023} &\underline{0.661} \\
CUT3R~\cite{wang2025continuous} &\underline{0.213} &\underline{0.066} &\textbf{0.621} &\underline{0.046} &\underline{0.015} &\textbf{0.473} &\underline{0.099} &\textbf{0.022} &\textbf{0.600} \\
\textbf{\sexyname{} (Ours)} &\textbf{0.189} &\textbf{0.054} &\underline{0.694} &0.092 &\textbf{0.012} &1.106 &0.176 &0.028 &1.204 \\
\bottomrule
\end{tabular}
\vspace{-1.5em}
\end{table*}

In this section, we demonstrate that \sexyname{} can achieve zero-shot reconstruction metrics comparable to or even better than SOTA reconstruction methods. We mainly consider two zero-shot reconstruction tasks: video depth estimation and camera pose estimation. Note that we only denoise for 4 steps for reconstruction tasks. 

\subsection{Zero-Shot Video Depth Estimation}
\label{sec:recon_exp-video-depth}

\noindent\textbf{Implementation Details.}
Video depth estimation is evaluated based on two key aspects: per-frame depth quality and inter-frame depth consistency. These evaluations are performed by aligning the predicted depth maps with the ground truth using a per-sequence scale. We use absolute relative error (Abs Rel) and $\delta < 1.25$ (percentage of predicted depths within a $1.25$-factor of true depth) as metrics. For implementation, we adopt the settings outlined in CUT3R~\cite{wang2025continuous}. Our baselines include both reconstruction-based methods—such as DUSt3R~\cite{wang2024dust3r}, MASt3R~\cite{leroy2024grounding}, MonST3R~\cite{zhang2024monst3r}, Spann3R~\cite{wang20243d}, and CUT3R~\cite{wang2025continuous}—and diffusion-based depth estimators, including ChronoDepth~\cite{shao2024learning}, DepthCrafter~\cite{hu2024depthcrafter}, and DepthAnyVideo (DA-V)~\cite{yang2024depth}. It is important to note that when comparing with diffusion-based depth estimators, we apply scale and shift alignment to the ground truth, as most of these methods are not inherently scale-invariant. All videos are resized with original aspect ratios kept to make the short side align with our model's input size.
For videos that exceed the maximum forward processing spatial or temporal size of our model, we employ a sliding window strategy with a stride size of $8$. In regions of overlap between windows, we first estimate a relative scale by calculating the average of element-wise division. This relative scale is then used to adjust the latter window's depth predictions. Finally, a linspace-weighted average is applied to the overlapping areas, following approaches similar to prior methods~\cite{hu2024depthcrafter, yu2022monosdf}.

\noindent\textbf{Results and Analysis.} 
Table~\ref{tab:video-depth} summarizes the video depth estimation results across Sintel~\cite{butler2012naturalistic}, BONN~\cite{palazzolo2019refusion}, and KITTI~\cite{geiger2013vision} datasets. For reconstruction-based methods, \sexyname{} outperforms or is comparable with prior approaches. On Sintel, \sexyname{} achieves the lowest Abs Rel (0.324), surpassing MonST3R-GA (0.378), and competitive $\delta<1.25$ (50.2). On KITTI, \sexyname{} sets a new benchmark with Abs Rel of 0.056 and $\delta<1.25$ of 97.8, outperforming the previous SOTA CUT3R (Abs Rel: 0.118, $\delta<1.25$: 88.1).
Among diffusion-based methods, \sexyname{} shows consistent superiority. It achieves the best performance on Sintel (Abs Rel: 0.314, $\delta<1.25$: 60.4) and KITTI (Abs Rel: 0.054, $\delta<1.25$: 97.7), significantly outperforming ChronoDepth~\cite{shao2024learning}, DepthCrafter~\cite{hu2024depthcrafter}, and DA-V~\cite{yang2024depth}. On BONN, \sexyname{} achieves the highest $\delta<1.25$ (60.2) with competitive Abs Rel (0.308).

\subsection{Zero-Shot Camera Pose Estimation}
\label{sec:recon_exp-camera-pose}

\noindent\textbf{Implementation Details.}
Following MonST3R~\cite{zhang2024monst3r} and CUT3R~\cite{wang2025continuous}, we evaluate camera pose estimation accuracy on the Sintel~\cite{butler2012naturalistic}, TUM Dynamics~\cite{sturm2012benchmark}, and ScanNet~\cite{dai2017scannet} datasets. Notably, both Sintel and TUM Dynamics contain highly dynamic objects, presenting significant challenges for traditional Structure-from-Motion (SfM) and Simultaneous Localization and Mapping (SLAM) systems. We report Absolute Translation Error (ATE), Relative Translation Error (RPE Trans), and Relative Rotation Error (RPE Rot) after Sim(3) alignment with the ground truth, following the methodology in~\cite{wang2025continuous}. The implementation settings are consistent with those used in CUT3R~\cite{wang2025continuous}. All videos are resized with original aspect ratios kept and then center cropped to align with our model's input size. For long videos exceeding our model's maximum temporal forward processing length, a sliding window strategy with a stride size of $32$ is employed. In overlapping regions between windows, camera poses are aligned following prior methods \cite{wang2024vggsfm}. Translation alignment is performed using linear interpolation, while quaternion rotations are interpolated with spherical linear interpolation. Additionally, we observed that the generated camera trajectories exhibit noise, likely due to the limited number of denoising steps. To mitigate this, we apply a simple Kalman filter~\cite{welch1995introduction} to smooth the trajectories.

\noindent\textbf{Results and Analysis.}
Table~\ref{tab:camera-pose} shows the evaluation results. Among feed-forward methods, \sexyname{} achieves the best ATE (0.189) and RPE Trans (0.054) on Sintel~\cite{butler2012naturalistic}, while remaining competitive in RPE Rot (0.694) compared to CUT3R (0.621). On TUM Dynamics~\cite{sturm2012benchmark}, \sexyname{} achieves the best RPE Trans (0.012). For other metrics, \sexyname{} is also comparable with other specialist models.

\begin{table*}[!tp]\centering
\caption{\textbf{VBench~\cite{huang2024vbench} Metrics of Video Prediction without Action Conditions.} Comparison between CogVideoX and \sexyname{} (Ours) on \textbf{\textit{in-domain/out-domain/overall}} performance on the validation set. For each group, the better performance is highlighted in \textbf{bold}.}\label{tab:i2v-wo-action}
\vspace{-1.0em}
\tablestyle{2.25pt}{0.95}
\begin{tabular}{l|cccccc|c}\toprule
&subject consistency &b.g. consistency &motion smoothness &dynamic degree &aesthetic quality &imaging quality &weighted average \\\midrule
CogVideoX &89.36/84.61/87.77 &92.72/91.43/92.29 &98.24/96.93/97.81 &88.75/95.00/90.83 &\textbf{54.49}/\textbf{53.58}/\textbf{54.18} &55.38/52.29/54.35 &79.01/77.52/78.51 \\
\sexyname{} &\textbf{91.50}/\textbf{87.55}/\textbf{90.18} &\textbf{94.29}/\textbf{93.62}/\textbf{94.07} &\textbf{98.54}/\textbf{98.19}/\textbf{98.42} &\textbf{96.25}/\textbf{100.00}/\textbf{97.50} &54.36/52.58/53.77 &\textbf{55.08}/\textbf{54.88}/\textbf{55.01} &\textbf{80.34}/\textbf{79.42}/\textbf{80.04} \\
\bottomrule
\end{tabular}
\vspace{-1em}
\end{table*}

\begin{table*}[!tp]\centering
\caption{\textbf{VBench~\cite{huang2024vbench} Metrics of Action-Conditioned Video Prediction.} Comparison between CogVideoX and \sexyname{} (Ours) on \textbf{\textit{in-domain/out-domain/overall}} performance on the validation set. For each metric group, the better performance is highlighted in \textbf{bold}.}\label{tab:i2v-w-action}
\vspace{-1.0em}
\tablestyle{2.25pt}{0.95}
\begin{tabular}{l|cccccc|c}\toprule
&subject consistency &b.g. consistency &motion smoothness &dynamic degree &aesthetic quality &imaging quality &weighted average \\\midrule
CogVideoX &\textbf{91.56}/88.23/90.51 &92.98/92.29/92.77 &98.44/97.81/98.24 &83.87/\textbf{93.02}/86.76 &\textbf{56.19}/57.43/\textbf{56.58} &\textbf{56.48}/61.60/58.10 &79.56/80.70/79.92 \\
\sexyname{} &90.73/\textbf{93.27}/\textbf{91.54} &\textbf{93.61}/\textbf{95.03}/\textbf{94.06} &\textbf{98.53}/\textbf{98.62}/\textbf{98.56} &\textbf{100.00}/83.72/\textbf{94.85} &55.04/\textbf{56.50}/55.50 &53.89/\textbf{63.23}/\textbf{56.84} &\textbf{80.33}/\textbf{81.55}/\textbf{80.71} \\
\bottomrule
\end{tabular}
\vspace{-1em}
\end{table*}

\section{Generation and Planning Experiments}
\label{sec:gen_exp}

In this section, we first show video prediction, with or without action conditioning, quantitatively or qualitatively, in \secref{sec:gen_exp-video-pred}. We then show visual planning abilities in \secref{sec:gen_exp-visual-planning}. More visualizations are in the supplementary material.

\subsection{Video Prediction}
\label{sec:gen_exp-video-pred}
\noindent\textbf{Implementation Details.} 
We use CogVideoX-5b-I2V~\cite{yang2024cogvideox} as our baseline. To ensure a fair comparison, we construct a validation dataset comprising two subsets: in-domain and out-domain data. The in-domain subset includes novel, unseen scenes from the same synthetic environments as the training dataset, while the out-domain subset consists of data from entirely new synthetic environments. Both models are provided with the first frame as the observation image.
For action-free prediction, since CogVideoX depends heavily on text prompts, we use GPT-4o~\cite{hurst2024gpt} to generate image descriptions and predictions of future scenes as prompts for CogVideoX. In contrast, \sexyname{} is evaluated using empty text prompts. For action-conditioned prediction, we also labeled camera trajectories in the validation dataset and generated corresponding raymap sequences as action conditions for \sexyname{}. For the baseline, in addition to the prompts used for action-free prediction, we use GPT-4o~\cite{hurst2024gpt} to generate detailed descriptions of object and camera movements, enabling the baseline to use language as action conditions.
We use the default classifier-free guidance value of 6 on text prompts for CogVideoX and a value of 3 on the observation image for \sexyname{}. No classifier-free guidance is applied to action conditions to ensure fairness. Evaluation metrics follow VBench~\cite{huang2024vbench}, a standard benchmark for video generation, with additional details on prompts and evaluation metrics provided in the supplementary material.

\noindent\textbf{Image-to-Video Prediction.}
We first evaluate image-to-video prediction without action conditions. The results, presented in \tabref{tab:i2v-wo-action}, show that \sexyname{} consistently outperforms the baseline on both in-domain and out-domain validation sets. Notably, \sexyname{} demonstrates a larger performance improvement on out-domain data, which can likely be attributed to the baseline model’s pre-training data containing domains similar to the in-domain dataset.

\noindent\textbf{Action-Conditioned Video Prediction.}
To assess the effectiveness of our post-training in improving action control and action-following capabilities, we conduct action-conditioned video prediction experiments. The results, shown in \tabref{tab:i2v-w-action}, indicate that \sexyname{} consistently outperforms the baseline in both in-domain and out-domain settings. Notably, CogVideoX tends to generate static scenes with high visual and aesthetic quality, while \sexyname{} accurately follows the action conditions, producing highly dynamic scenes. These results validate the effectiveness of our framework and the advantages of using camera trajectories as action conditions. %

\begin{table*}[!tp]\centering
\caption{\textbf{Pixel-wise Metrics of Action-Conditioned Navigation.} Comparison of performance between \sexyname{}-no-depth and \sexyname{} on \textbf{\textit{in-domain/out-domain/overall}} performance. For each metric group, the better performance is highlighted in \textbf{bold}.}\label{tab:nav-w-action}
\vspace{-1.0em}
\tablestyle{16.5pt}{0.95}
\begin{tabular}{l|cccc}\toprule
&PSNR~$\uparrow$&SSIM~$\uparrow$ &MS-SSIM~$\uparrow$ &LPIPS~$\downarrow$ \\\midrule
\sexyname{}-no-depth &19.13/18.67/18.97 &0.5630/0.4830/0.5353 &0.5467/0.5204/0.5376 &0.3116/0.2995/0.3074 \\
\sexyname{} &\textbf{19.87}/\textbf{19.37}/\textbf{19.70} &\textbf{0.5803}/\textbf{0.5058}/\textbf{0.5545} &\textbf{0.5830}/\textbf{0.5627}/\textbf{0.5760} &\textbf{0.2691}/\textbf{0.2599}/\textbf{0.2659} \\
\bottomrule
\end{tabular}
\vspace{-1em}
\end{table*}

\begin{table*}[!tp]\centering
\caption{\textbf{Quantitative Results of Action-Free Visual Path Planning.} Comparison of performance between Aether and Aether-no-depth on \textbf{\textit{in-domain/out-domain/overall}} performance. For each metric group, the better performance is highlighted in \textbf{bold}.}\label{tab:nav-wo-action}
\vspace{-1.0em}
\tablestyle{2.25pt}{0.95}
\begin{tabular}{l|cccccc|c}\toprule
&subject consistency &b.g. consistency &motion smoothness &dynamic degree &aesthetic quality &imaging quality &weighted average \\\midrule
Aether-no-depth &88.68/89.61/88.61 &93.62/93.92/93.66 &98.37/98.31/98.32 &\textbf{97.06}/91.67/\textbf{96.15} &54.12/56.26/54.78 &51.77/58.46/54.29 &79.11/80.43/79.59 \\
Aether (Ours) &\textbf{89.69}/\textbf{91.61}/\textbf{90.36} &\textbf{93.88}/\textbf{94.58}/\textbf{94.13} &\textbf{98.50}/\textbf{98.40}/\textbf{98.46} &\textbf{97.06}/\textbf{91.67}/95.19 &\textbf{55.83}/\textbf{56.87}/\textbf{56.19} &\textbf{54.71}/\textbf{61.13}/\textbf{56.93} &\textbf{80.21}/\textbf{81.53}/\textbf{80.67} \\
\bottomrule
\end{tabular}
\vspace{-1.0em}
\end{table*}

\subsection{Visual Planning}
\label{sec:gen_exp-visual-planning}

\noindent\textbf{Implementation Details.} 
We evaluate the action-conditioned navigation capability of \sexyname{} on our validation set. To demonstrate the effectiveness of our multi-task objective, particularly the incorporation of the reconstruction objective, we also post-train an ablation model without the video depth objective, denoted as \sexyname{}-no-depth. Given the observation image, goal image, and camera trajectory, the resulting video should be highly determined. Thus, we report pixel-wise reconstruction metrics, including PSNR, SSIM~\cite{wang2004image}, MS-SSIM~\cite{wang2003multiscale}, and LPIPS~\cite{zhang2018unreasonable}, for action-conditioned navigation. For the action-free case, which represents a visual path navigation task, we also report the VBench metrics. We do \textit{not} use any classifier-free guidance on both tasks.

\noindent\textbf{Action-Conditioned Navigation.}
The quantitative results for action-conditioned navigation are presented in \tabref{tab:nav-w-action}. \sexyname{} consistently outperforms the ablation model, demonstrating the significant benefits of incorporating the reconstruction objective into generative models.

\noindent\textbf{Visual Path Planning.}
In the absence of action conditions, this task evaluates the model’s ability to function as a ``world model as an agent," requiring it to plan a path from the observation image to the goal image. The results, shown in \tabref{tab:nav-wo-action}, indicate that the reconstruction objective substantially improves the model's visual path planning capability. Additionally, qualitative visualizations on completely in-the-wild data are provided in supplementary material.

\section{Related Work}
\label{sec:related_work}

\noindent\textbf{World Models.}
World models have emerged as a critical framework in artificial intelligence, enabling agents to simulate, understand, and predict environmental dynamics. Early work~\cite{ha2018world} introduced latent representations and recurrent neural networks for decision-making. Recent advancements include Cat3D~\cite{gao2024cat3d} for 3D scene generation, Cat4D~\cite{wu2024cat4d} for dynamic 4D environments, and Genie 2~\cite{parkerholder2024genie2}, a large-scale model for interactive 3D worlds. Motion Prompting~\cite{geng2024motionpromptingcontrollingvideo} further enables precise video generation control. These advancements demonstrate the evolution of world models toward dynamic, interactive, and controllable applications in robotics, gaming, and simulation.

\noindent\textbf{Reconstruction.}
Reconstruction has been a long-standing topic in computer vision, with notable progress in both traditional and learning-based methods. Classical approaches, such as Structure-from-Motion (SfM) \cite{hartley2003multiple,cui2017hsfm,schonberger2016structure,pan2024global} and Multi-View Stereo (MVS) \cite{furukawa2015multi,schonberger2016pixelwise}, rely on multi-view geometry for feature matching, pose estimation, and dense point cloud generation, demonstrating robust performance in controlled settings. 
Deep learning has introduced powerful alternatives, tackling sub-tasks like feature matching \cite{sarlin2020superglue,edstedt2024roma}, point tracking \cite{doersch2022tap,wang2024vggsfm}, triangulation \cite{moran2021deep}, and MVS \cite{yao2018mvsnet,zhang2023vis}. End-to-end methods now directly predict point maps \cite{wang2024dust3r,leroy2024grounding} or depth maps from images \cite{yang2024depth1,bochkovskii2024depth}, often incorporating camera parameters \cite{wei2020deepsfm}. 
Recently, diffusion models have achieved breakthroughs in image and video generation \cite{ho2020denoising,nichol2021improved,yang2024cogvideox,kong2024hunyuanvideo,xu2025exploring}, inspiring novel 3D reconstruction approaches that leverage rich 2D priors \cite{ke2024repurposing,zhu2024spa,hu2024depthcrafter,zhu2023ponderv2,yang2024depth,fu2024geowizard,yang2024unipad,lu2025matrix3d}. These methods demonstrate the potential of integrating diffusion-based 2D knowledge into 3D modeling.

\noindent\textbf{Video Generation.}
Video generation has evolved from foundational techniques like DDPM~\cite{ho2020denoising,nichol2021improved} to modern frameworks leveraging diffusion-based techniques. Advances such as latent diffusion~\cite{rombach2022high} and diffusion transformers~\cite{peebles2023scalable} have improved generation quality, while models like Sora~\cite{videoworldsimulators2024} and Stable Video Diffusion (SVD)~\cite{blattmann2023stable} emphasize temporal consistency. Open-source models, including LTX Video~\cite{HaCohen2024LTXVideo}, CogVideoX~\cite{yang2024cogvideox}, and Hunyuan Video~\cite{kong2024hunyuanvideo}, offer increased flexibility, and techniques like multi-scale architectures (e.g., Pyramid Flow~\cite{jin2024pyramidal}) enhance motion dynamics. These advancements highlight rapid progress, with ongoing efforts to improve scalability and temporal stability.

\section{Conclusion and Limitations}
\label{sec:conclusion}
In this work, we introduce \sexyname{}, a geometry-aware multi-task world model that reconstructs 4D dynamic videos, predicts future frames conditioned on observation images and actions, and performs visual planning based on observation and goal images. We propose an automatic 4D synthetic data labeling pipeline, enabling \sexyname{} to train on synthetic data and generalize to unseen real-world data in a zero-shot manner. Post-trained on the CogVideoX base model, \sexyname{} achieves state-of-the-art or competitive reconstruction performance and outperforms baselines in generation and planning tasks, demonstrating the value of incorporating reconstruction objectives into world modeling.

However, limitations remain. Camera pose estimation is less accurate, likely due to incompatibilities between raymap representation and prior video diffusion models. Indoor scene reconstruction also lags behind outdoor performance, likely due to the predominance of outdoor training data. Additionally, predictions without language prompts often fail in highly dynamic scenes. Future work can address these by exploring novel action representations, co-training with real-world data, and retaining the language prompting capabilities of the base model.

\newpage
\section*{Acknowledgments}
This work is supported by the National Key R\&D Program of China (NO.2022ZD0160102) and Shanghai Artificial Intelligence Laboratory.

{
    \small
    \bibliographystyle{ieeenat_fullname}
    \bibliography{main}
}

\newpage
\appendix
\newpage
\appendix

\section{Author Contributions}
All authors contributed equally.

\begin{itemize}
    \setlength{\itemsep}{5pt}
    \setlength{\parsep}{5pt}
    \setlength{\parskip}{5pt}
    \item \textbf{Network Architecture and Model Training} \\
        Haoyi Zhu (Shanghai AI Lab, USTC), \\
        Junyi Chen (Shanghai AI Lab, SJTU),
    \item \textbf{Data Collection and Automatic Labeling Pipeline} \\
        Yifan Wang (Shanghai AI Lab, SJTU), \\
        Jianjun Zhou (ZJU, Shanghai AI Lab, SII), \\
        Wenzhang Chang (Shanghai AI Lab, USTC), \\
        Zizun Li (Shanghai AI Lab, USTC), \\
        Yang Zhou (Shanghai AI Lab, FDU), 
    \item \textbf{Model Evaluation} \\
        Haoyi Zhu (Shanghai AI Lab, USTC), \\
        Wenzheng Chang (Shanghai AI Lab, USTC),
    \item \textbf{Paper (figures, visualizations, writing)} \\
        Haoyi Zhu (Shanghai AI Lab, USTC), \\
        Wenzheng Chang (Shanghai AI Lab, USTC), \\
        Junyi Chen (Shanghai AI Lab, SJTU), \\
        Jianjun Zhou (ZJU, Shanghai AI Lab, SII), \\
        Yifan Wang (Shanghai AI Lab, SJTU), \\
        Tong He (Shanghai AI Lab),
    \item \textbf{Leadership (managed and advised on the project)} \\
        Tong He (Shanghai AI Lab),
    \item \textbf{Consultant (provided valuable advice)} \\
        Chunhua Shen (ZJU), \\
        Jiangmiao Pang (Shanghai AI Lab)\\
\end{itemize}

\noindent We also want to thank Mingyu Liu and Kaipeng Zhang for the helpful discussion. 

\section{Robustness of Data Annotation Pipeline}
Here we detail three key design choices in our methodology that were specifically implemented to enhance its robustness against common sources of uncertainty in dynamic RGB-D processing.

\noindent\textbf{Robustness in Dynamic Masking}
Grounding SAM 2 often yields erroneous results for out-of-domain semantic inputs. To enhance the robustness of this process, we select prompts with low uncertainty and discard frames with a high mask-to-image ratio. This approach improves the reliability of our dynamic mask generation, thereby increasing the robustness of all subsequent operations.

\noindent\textbf{Robustness Against Inaccurate Flow Estimation}
In our video slicing process, we utilize optical flow magnitude and the forward-backward error as key metrics. This approach mitigates the uncertainty inherent in flow estimation during coarse camera pose estimation, leading to more robust initial annotations.

\noindent\textbf{Robustness in Points Trajectory Estimation}
Similarly, our video slicing is performed based on optical flow magnitude and forward-backward error. In addition, we discard frames with an insufficient number of keypoints. These steps yield a video sequence that is both rich in keypoints and temporally coherent (i.e., without frame discontinuities). Such a sequence is highly conducive to tracking estimation methods and these operations also serve to minimize the uncertainty associated with the tracking process.

\noindent\textbf{Robustness in Failure Sequence Filtering}
As a final step, we filter out erroneous estimations using three key criteria. We discard an entire sequence if it exhibits an anomalous focal length, if its reprojection error relative to point tracking exceeds a predefined threshold, or if its geometric consistency error surpasses a specified limit.

\noindent\textbf{Conclusion on Overall Robustness}
Our method consistently yields accurate and clean camera poses with minimal noise. Furthermore, the safeguarding operations detailed above ensure that our processed data is virtually free of failure cases. This outcome is the key to the robustness of our approach.

\section{Raymap to Camera Parameters Algorithm}

We adopt a direct approach to recover camera parameters from raymaps, as shown in Algorithm~\ref{alg:raymap-to-cam-pose}. For more details, please refer to our \href{https://github.com/OpenRobotLab/Aether}{GitHub repository}.

\begin{algorithm}[ht]
\caption{Raymap to camera parameters conversion.}
\label{alg:raymap-to-cam-pose}
\algcomment{\fontsize{7.2pt}{0em}\selectfont \texttt{normalize}: L2 normalization; \texttt{cross}: cross product; \texttt{eye}: identity matrix.
}
\definecolor{codeblue}{rgb}{0.25,0.5,0.5}
\lstset{
  backgroundcolor=\color{white},
  basicstyle=\fontsize{7.2pt}{7.2pt}\ttfamily\selectfont,
  columns=fullflexible,
  breaklines=true,
  captionpos=b,
  commentstyle=\fontsize{7.2pt}{7.2pt}\color{codeblue},
  keywordstyle=\fontsize{7.2pt}{7.2pt},
}

\begin{lstlisting}[language=python]
# Inputs: ray_o (N,H,W,3), ray_d (N,H,W,3)
# Outputs: extrinsics (N,4,4), intrinsic (N,3,3)

# 1. Estimate Camera Position and Orientation
# ------------------------------------------
c = mean(ray_o.reshape(N,-1,3), dim=1)  # camera center

# Look-at point is average of ray endpoints
p = mean((ray_o + ray_d).reshape(N,-1,3), dim=1)

# Camera coordinate frame
z = normalize(p - c)  # Forward axis (N,3)
x = normalize(mean(ray_d[:,:,-1], dim=1) - mean(ray_d[:,:,0], dim=1))  # Right axis (N,3)
y = normalize(cross(z, x))  # Up axis (N,3)
x = normalize(cross(y, z))  # Ensure orthogonality

# 2. Construct Poses Matrix
# -----------------------------
R = stack([x, y, z], dim=2)  # Rotation (N,3,3)
t = c.unsqueeze(-1)  # Translation (N,3,1)
poses = eye(4).repeat(N,1,1)
poses[:,:3,:3] = R
poses[:,:3,3] = c

# 3. Construct Intrinsics Matrix
# ----------------------------
intrinsics = eye(3).repeat(N,1,1)
intrinsics[:,0,0] = norm(p - c) # Focal length
intrinsics[:,1,1] = norm(p - c) # Assume fx = fy
intrinsics[:,0,2] = W / 2   # Principal point
intrinsics[:,1,2] = H / 2   # Assume at center

extrinsics = inverse(poses)

return extrinsics, intrinsics
\end{lstlisting}
\end{algorithm}

\section{Generation Experiments Details}

\paragraph{Prediction Validation Dataset Construction.}
For the validation set of prediction tasks, we collected 93 in-domain scenes and 43 out-of-domain scenes, with each scene corresponding to a synthetic video clip. The in-domain scenes are collected from the same synthetic environments used in the training dataset, while the out-of-domain scenes are sourced from entirely different synthetic environments that are not present in the training data.
\paragraph{Video Prediction Task Settings.}

For prediction tasks without action conditions, both Aether and CogVideoX take the first frame as input. However, since CogVideoX tends to generate static scenes without text prompts, we utilize GPT-4o to generate text annotations for each image. The prompt for GPT-4o is designed to:

1) Generate text labels describing the scene content
2) Predict the potential motion patterns of each object
3) Predict the most likely camera trajectory based on the image content
4) For scenes with clear subjects, predict camera movements that follow the subject
5) For scenes without prominent subjects, predict reasonable camera movements based on the scene context
6) Emphasize dynamic video generation with camera movements that closely track subjects or rapidly move to showcase the scene

The generated text labels and the first frame serve as input for CogVideoX, with a negative prompt set to ``static background, static camera, slow motion, slow camera movement, low dynamic degree" and a guidance scale of 6.0. In contrast, Aether only takes the first frame as input, and sets obs guidance scale to 3.0. 

\paragraph{Action Conditioned Video Prediction Task Settings.}

For action conditioned prediction tasks, Aether accepts both the first frame as observation image input and the camera trajectory of the video clip as action-conditioned input. To ensure fair comparison, we use GPT-4o to generate detailed text annotations for both the initial and final frames of each video clip. These annotations serve as text prompts for CogVideoX, providing comprehensive camera trajectory descriptions. The prompt template for GPT-4o is designed to:

1) Describe the initial frame in detail
2) Predict the video content based on both frames, including:
   - Object movements and interactions
   - Scene dynamics
   - Camera motion patterns
3) Analyze the differences between the start and end frames to:
   - Determine the precise camera movement trajectory
   - For scenes with clear subjects, describe how the camera follows them
   - For scenes without prominent subjects, predict the most probable camera movements
4) Emphasize dynamic scene generation with active camera movements

This approach provides CogVideoX with more detailed camera motion descriptions compared to the action-free setting, serving as an equivalent to Aether's explicit action conditions.

\paragraph{VBench Evaluation Protocol.}

We adopt VBench as our evaluation metric system for prediction tasks. Given the differences in input settings between Aether and CogVideoX, we evaluate the generated videos under the custom input configuration of VBench across six dimensions:

1) Subject Consistency: Evaluates the temporal consistency of main subjects

2) Background Consistency: Measures the stability and coherence of scene backgrounds

3) Motion Smoothness: Assesses the fluidity and naturalness of movements

4) Dynamic Degree: Quantifies the level of motion and activity

5) Aesthetic Quality: Measures the visual appeal and artistic merit

6) Imaging Quality: Evaluates the technical quality of video generation

The final score is computed as a weighted average of these dimensions using the official VBench weights:
\begin{itemize}
    \item Subject Consistency: 1.0
    \item Background Consistency: 1.0
    \item Motion Smoothness: 1.0
    \item Dynamic Degree: 0.5
    \item Aesthetic Quality: 1.0
    \item Imaging Quality: 1.0
\end{itemize}

Based on the VBench evaluation results, as shown in Tables~\ref{tab:i2v-wo-action} and~\ref{tab:i2v-w-action}, Aether demonstrates superior overall performance compared to CogVideoX across these metrics.

\paragraph{Video Planning Settings.}
For planning tasks, we construct a validation set following a similar approach to the prediction tasks, comprising 80 in-domain scenes and 40 out-of-domain scenes from synthetic environments. For each video clip, we extract the initial and final frames as inputs for both Aether and Aether-no-depth models.

For action-conditioned tasks, we evaluate model performance using pixel-wise metrics (PSNR, SSIM, MS-SSIM, and LPIPS) as shown in Table~\ref{tab:nav-w-action}. For action-free tasks, we employ the VBench evaluation metrics as presented in Table~\ref{tab:nav-wo-action}. Both evaluation protocols demonstrate that Aether consistently outperforms the Aether-no-depth model, validating the effectiveness of our approach.

\section{Additional Losses in Stage 2 Training}

In our second training stage, we decode the latent representations into image space and employ three distinct losses: MS-SSIM loss for color videos, Scale- and Shift-Invariant (SSI) loss for depth videos, and Pointmap loss for raymaps. Each loss is tailored to the unique characteristics of the respective modality, ensuring effective supervision across all tasks.

\subsection{Multi-Scale Structural Similarity (MS-SSIM) Loss for Color Videos} 
For color videos, we use the Multi-Scale Structural Similarity (MS-SSIM) loss to preserve perceptual quality and structural coherence across multiple scales. Unlike pixel-wise losses, MS-SSIM captures luminance, contrast, and structural differences between predicted \( \mathbf{\hat{I}} \) and ground truth \( \mathbf{I} \) frames. At each scale, the structural similarity index is computed as:
\[
\text{SSIM}(\mathbf{\hat{I}}, \mathbf{I}) = \frac{(2\mu_{\hat{I}}\mu_I + C_1)(2\sigma_{\hat{I}I} + C_2)}{(\mu_{\hat{I}}^2 + \mu_I^2 + C_1)(\sigma_{\hat{I}}^2 + \sigma_I^2 + C_2)},
\]
where \( \mu_{\hat{I}}, \mu_I \) are local means, \( \sigma_{\hat{I}}, \sigma_I \) are standard deviations, \( \sigma_{\hat{I}I} \) is the cross-covariance, and \( C_1, C_2 \) are constants to stabilize division. MS-SSIM is computed across multiple scales by downsampling the input, with weights \( \{w_i\} \):
\[
\text{MS-SSIM} = \prod_{i=1}^M \text{SSIM}_i^{w_i}.
\]
The MS-SSIM loss is defined as:
\[
\mathcal{L}_\text{MS-SSIM} = 1 - \text{MS-SSIM}.
\]
This loss is particularly effective for color videos, as it emphasizes structural similarity over pixel-wise accuracy.

\subsection{Scale- and Shift-Invariant (SSI) Loss for Depth Videos}
Depth predictions often suffer from scale and shift ambiguities. To address this, we use a Scale- and Shift-Invariant (SSI) loss, which aligns the predicted depth \( \mathbf{\hat{D}} \) with the ground truth \( \mathbf{D} \) by computing optimal scale \( s \) and shift \( t \) as follows:
\[
s, t = \arg\min_{s, t} \| \mathbf{M} \odot (s \mathbf{\hat{D}} + t - \mathbf{D}) \|^2,
\]
where \( \mathbf{M} \) is a binary mask for valid pixels, and \( \odot \) is the element-wise product. The SSI loss combines a data term and a gradient regularization term:
\[
\mathcal{L}_\text{SSI} = \mathcal{L}_\text{data} + \alpha \mathcal{L}_\text{gradient},
\]
where \( \alpha \) balances the contribution of gradient regularization. The gradient term enforces local smoothness in depth predictions, ensuring geometric consistency.

\subsection{Pointmap Loss for Raymaps}
Raymaps encode 3D spatial information, and their alignment requires a loss invariant to scale and translation. We transform predicted disparity and raymaps into 3D pointmaps \( \mathbf{P} \) using:
\[
\mathbf{P} = \mathbf{D} \cdot \mathbf{R}_d + \mathbf{R}_o,
\]
where \( \mathbf{D} \) is the depth, \( \mathbf{R}_d \) is the ray direction, and \( \mathbf{R}_o \) is the ray origin. The pointmap loss minimizes the difference between predicted and ground truth pointmaps:
\[
\mathcal{L}_\text{pointmap} = \frac{1}{N} \sum_{i=1}^N w_i \| \mathbf{\hat{P}}_i - \mathbf{P}_i \|_p,
\]
where \( w_i \) is a weight inversely proportional to depth, \( p \) is the norm type (e.g., \( L_1 \) or \( L_2 \)), and \( N \) is the number of valid points. This loss ensures accurate 3D spatial alignment, which is critical for raymap-based tasks. Note that the pointmap loss only back-propagates gradients to raymap latents, and we stop the disparity gradients during pointmap projection.

\begin{figure*}[!t]
    \centering
    \includegraphics[width=0.95\linewidth]{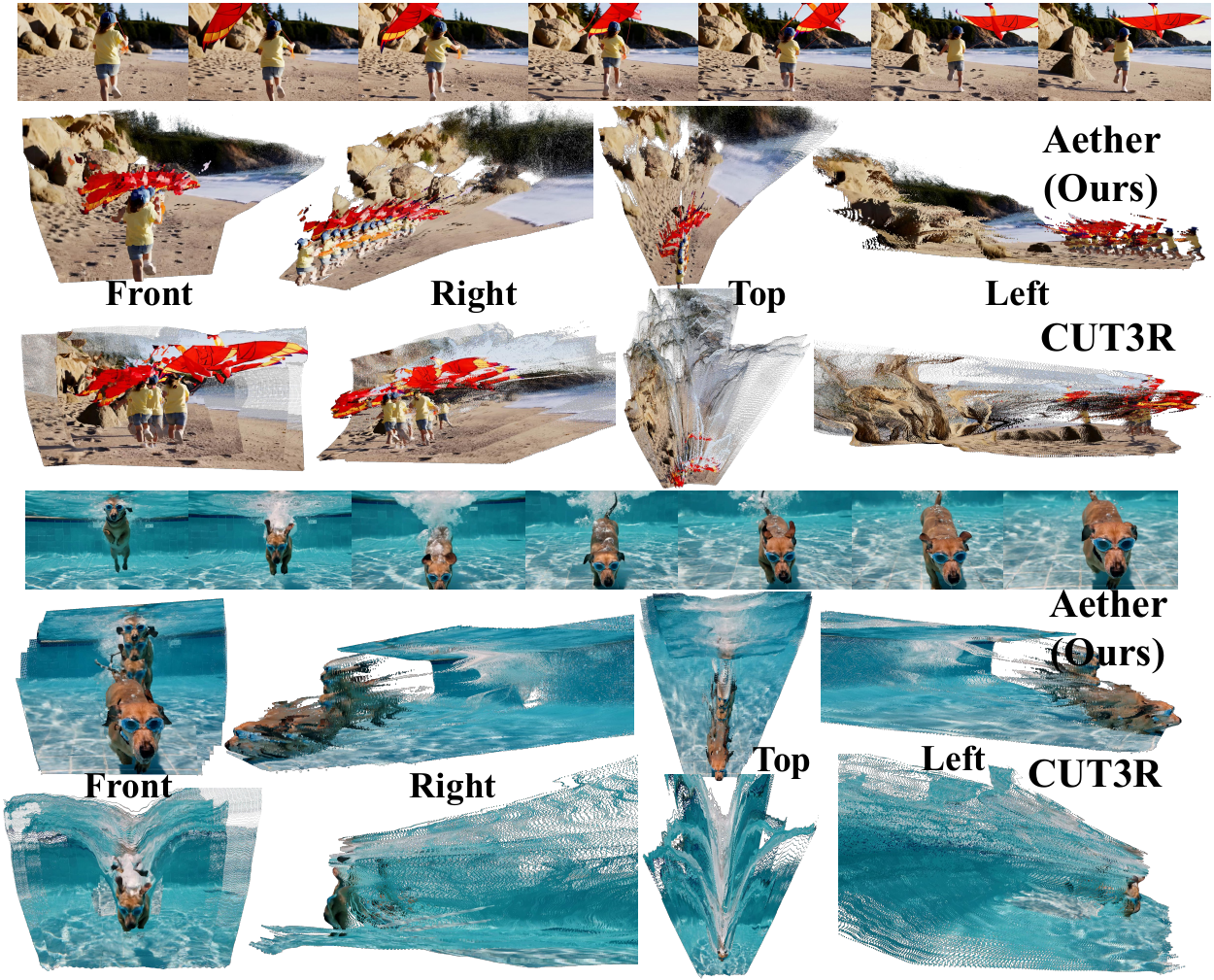}
    \caption{More reconstruction visualizations.}
    \label{fig:recon-vis}
\end{figure*}

\begin{figure*}[!t]
    \centering
    \includegraphics[width=0.95\linewidth]{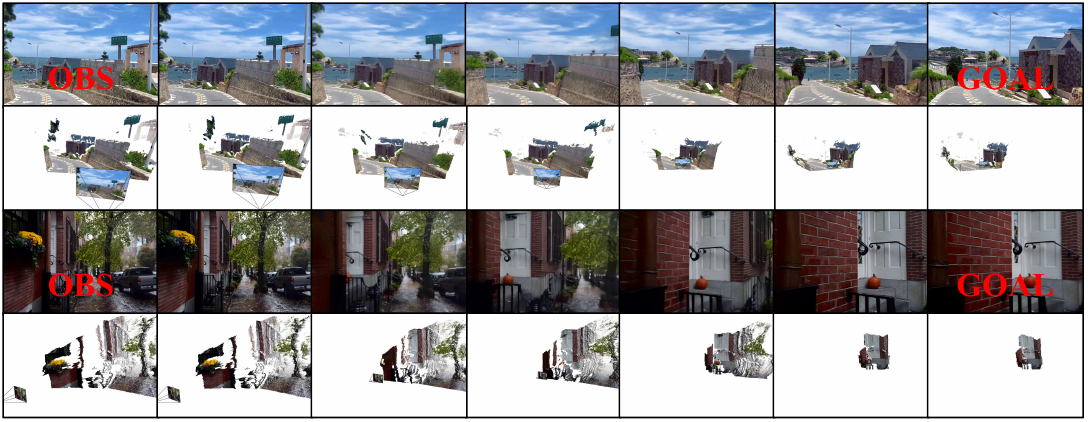}
    \caption{More visual planning examples.}
    \label{fig:planning-vis}
\end{figure*}

\begin{figure*}[!t]
    \centering
    \includegraphics[width=0.95\linewidth]{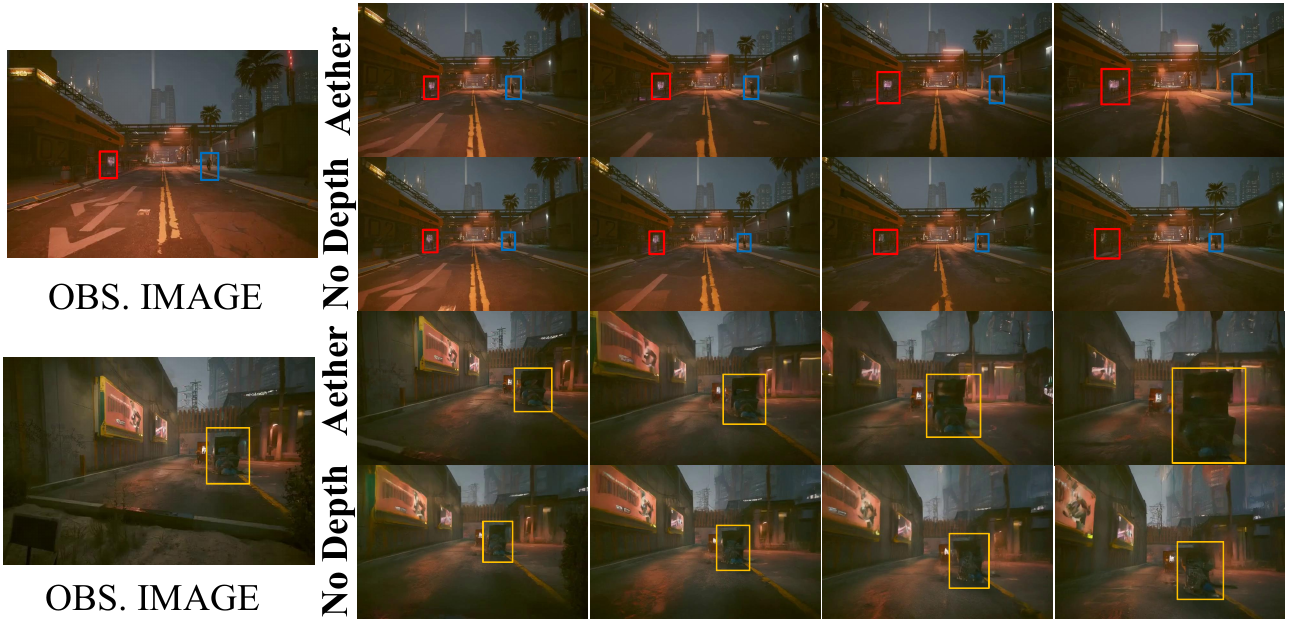}
    \caption{Qualitative results for ablation study. \textbf{\textit{Please zoom in.}}}
    \label{fig:depth-ablation}
\end{figure*}

\section{More Ablation Study}

Acknowledging the importance of ablation studies and working within our computational resources, we conducted a key ablation in Sec. 5.2, where the depth component was removed during training. Results presented in Tab. 5 and 6 demonstrate that excluding the 4D reconstruction target from the multi-task co-training leads to a notable degradation in visual planning performance. This finding strongly supports our paper's central claim regarding the effective integration of reconstruction and generation within a unified framework. Qualitative results further illustrating this are provided in Fig~\ref{fig:depth-ablation}.

\section{More analysis in Sec. 4} 

Our model performs well on the Sintel and Kitti datasets but is comparatively weaker on BONN. The trend is also observed in other diffusion-based methods. We suggest two primary reasons for this. First, BONN's scene type is indoor. This may be less compatible with the learned priors of video diffusion models. Second, as an older dataset, BONN exhibits lower image quality and contains artifacts such as motion blur. Diffusion models can be particularly sensitive to such image characteristics, potentially impacting their performance.

\section{More Training data details}
Our synthetic data collection approach directly follows DA-V and TheMatrix, capturing RGB-D videos from AAA games such as Cyberpunk2077 and Horizon5. The initial raw dataset contained about 12.5 million frames. After undergoing camera pose annotation and filtering, this collection was refined to approximately 8.9 million well-annotated frames, which were subsequently used for training. 
Our camera pose estimation is comparable to other feed-forward methods, which typically trade the higher accuracy of optimization-based techniques for superior run-time efficiency. Reduced performance on ScanNet is likely due to the domain gap from synthetic training data, alongside ScanNet's imperfect annotations and motion blur.

\section{Running time differences.}
See Tab.~\ref{tab:running-time}.

\begin{table}[!htp]\centering
\caption{Reconstruction running FPS differences on A100.}\label{tab:running-time}
\tablestyle{2pt}{0.95}
\resizebox{\columnwidth}!{
\begin{tabular}{l|ccccc}\toprule
Method &DUSt3R-GA &MASt3R-GA &MonST3R-GA &Aether (Ours) \\\cmidrule{1-5}
Resolution &144 $\times$ 512 &144 $\times$ 512 &144 $\times$ 512 &480 $\times$ 640 \\
FPS &0.76 &0.31 &0.35 &6.14 \\
\bottomrule
\end{tabular}
}
\end{table}

\end{document}